\documentclass[11pt,letterpaper, twocolumn]{article}
\usepackage[top=0.7in, bottom=0.7in, left=0.8in, right=0.8in]{geometry}
\usepackage{times}
\usepackage{latexsym}
\usepackage{graphicx}
\usepackage{balance}
\usepackage{hyperref}
\usepackage{natbib}
\usepackage{arydshln}
\usepackage{enumitem}
\usepackage{subcaption}

\newcommand{\squishlist}{
   \begin{list}{$\bullet$}
    { \setlength{\itemsep}{-1pt}      \setlength{\parsep}{3pt}
      \setlength{\topsep}{1pt}       \setlength{\partopsep}{0pt}
      \setlength{\leftmargin}{1.5em} \setlength{\labelwidth}{1em}
      \setlength{\labelsep}{0.5em} } }
\newcommand{\squishend}{
    \end{list}  }
    
\title{Closed-Loop Robotic Manipulation of Transparent Substrates for Self-Driving Laboratories using Deep Learning Micro-Error Correction} 
\author{}

\date{}

\begin{document}

\twocolumn[
\begin{@twocolumnfalse}
\maketitle

% Pull the author block up closer to the title:
\vspace{-4.0em}   % <-- adjust this value to taste

\begin{center}
  % Top row: 3 authors
  \begin{tabular}{ccc}
    \shortstack{Kelsey Fontenot\\MIT\\\texttt{kelfon@mit.edu}} &
    \shortstack{Anjali Gorti\\MIT\\\texttt{argorti@mit.edu}} &
    \shortstack{Iva Goel\\MIT\\\texttt{ivagoel@mit.edu}}
  \end{tabular}

  \vspace{0.8em} % space between the two rows of authors

  % Bottom row: 2 authors
  \begin{tabular}{cc}
    \shortstack{Tonio Buonassisi\\MIT\\\texttt{buonassisi@mit.edu}} &
    \shortstack{Alexander E. Siemenn\\MIT\\\texttt{asiemenn@mit.edu}}
  \end{tabular}
\end{center}

\vspace{1em} % space before the main text
\end{@twocolumnfalse}
]

\begin{abstract}
Self-driving laboratories (SDLs) have accelerated the throughput and automation capabilities for discovering and improving chemistries and materials. Although these SDLs have automated many of the steps required to conduct chemical and materials experiments, a commonly overlooked step in the automation pipeline is the handling and reloading of substrates used to transfer or deposit materials onto for downstream characterization. Here, we develop a closed-loop method of Automated Substrate Handling and Exchange (ASHE) using robotics, dual-actuated dispensers, and deep learning-driven computer vision to detect and correct errors in the manipulation of fragile and transparent substrates for SDLs. Using ASHE, we demonstrate a 98.5\% first-time placement accuracy across 130 independent trials of reloading transparent glass substrates into an SDL, where only two substrate misplacements occurred and were successfully detected as errors and automatically corrected. Through the development of more accurate and reliable methods for handling various types of substrates, we move toward an improvement in the automation capabilities of self-driving laboratories, furthering the acceleration of novel chemical and materials discoveries.
\end{abstract}

\section{Introduction}
In recent years, laboratory automation and self-driving laboratories (SDLs) have emerged as solutions to accelerate chemical and materials discovery research \cite{abolhasani2023rise, tobias2025autonomous, tom2024selfdriving}. Automated methods are essential for these discovery pipelines as they require precise repeatability, high-throughputs, and continuous operation to explore meaningful volumes of high-dimensional materials search spaces \cite{yik2023automated, nishio2025digital, szymanski2023autonomous, macleod2020selfdriving}, such as electrolytes \cite{sorkun2025redcat, crabtree2020selfdriving}, polymers \cite{knox2025selfdriving, wang2025autonomous}, and perovskites \cite{sadeghi2025selfdriving, xu2025autonomous}. SDLs and other robotic equipment are effective at automating time-consuming and repetitive tasks necessary for executing these experiments, while also improving task precision due to a coupled intelligent software-hardware control approach \cite{yik2023automated, soh2023automated, nishio2025digital}. On the software end, computer vision and machine learning models collect and learn from data to improve the accuracy and precision of tasks within the materials discovery pipeline \cite{pickles2025automated, zhang2024learning, lin2025visual, hickman2025atlas}. Then, on the hardware end, automated pipetters and robotic arms execute those commands with high precision using those software outputs \cite{yik2023automated, macleod2020selfdriving, szymanski2023autonomous}. 

\begin{figure*}[h]
\centering
\begin{subfigure}{2\columnwidth}  
\includegraphics[width=\columnwidth]{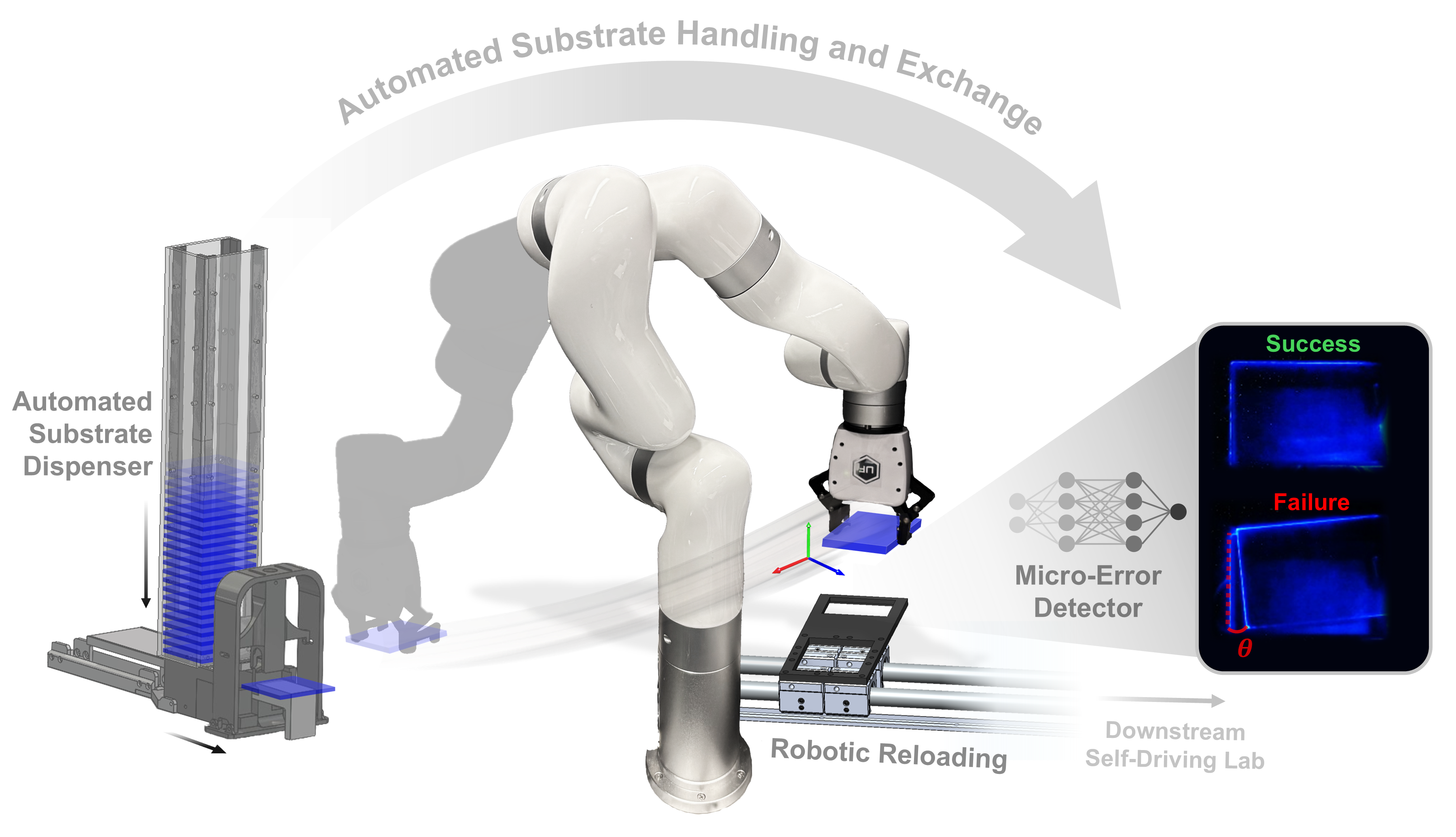 }
\end{subfigure}\hfill%
\caption{\footnotesize The Automated Substrate Handling and Exchange (ASHE) system for robotic manipulation and error correction of transparent substrate placement. ASHE combines the use of a dual-actuated dispenser, a 5-degree-of-freedom robotic arm with a specialized deformable gripper, and a deep learning computer vision detector to automate substrate reloading and unloading from experimental equipment. Through vision-based detection and automated correction of micro-scale errors in the placement of a substrate, ASHE carefully manipulates fragile and transparent substrates with high accuracy and repeatability, suited for the long, autonomous experimental campaigns of self-driving laboratories.}
\label{fig:ashe}
\end{figure*}

Robotic arms are widely used in SDLs for their reliability, precision, and speed\cite{bayley2024autonomous, burger2020mobile, coley2019robotic, macleod2020selfdriving}. For example, robotic arms have been used in literature to automate materials preparation, synthesis, and characterization, including powder X-ray diffraction (XRD) \cite{yotsumoto2024autonomous}, Reversible Addition–Fragmentation Chain-Transfer (RAFT) polymerization \cite{lee2023fully}, and solid dispensing \cite{jiang2023autonomous}. In combination with computer vision and fiducial markers, robotic arms use path planning to autonomously navigate to different positions for picking and placing of samples \cite{fernando2025robotic}. As part of a larger materials discovery pipeline, an important but often overlooked sub-task of an SDL is to replace a substrate that has been used to transfer and deposit materials onto for downstream characterization with a fresh, unused substrate for the next round of materials deposition \cite{macleod2020selfdriving, fushimi2025development}. Without automation, this replacement process must be conducted \textit{via} human intervention, bottlenecking the pipeline. Substrate types vary greatly across different materials and chemistry domains. For example, 96-well plates are used as substrates for aqueous pipetting of electrolytes or polymers \cite{matsuda2022data, bai2019accelerated}, metal alloyed plates are used as substrates for high-temperature sintered ceramic coatings \cite{lu2007constrained}, and silica or glass slides are used as substrates for conductive carbon nanotubes or semiconductor perovskites \cite{bash2021multi, um2025tailoring}. Robotic arms can be used to replace these substrates for the next cycle of an experiment. However, this task becomes significantly more challenging to perform autonomously with high accuracy and precision as the substrates become harder to detect and handle, \textit{i.e.}, transparent, thin, and fragile glass or silica substrates, as often used in experiments with conductive materials \cite{fraser1972highly, bash2021multi, ma2019graphene, um2025tailoring, siemenn2024using, sun2021data}.

\begin{figure*}[ht!]
\centering
\begin{subfigure}{2\columnwidth}  
\includegraphics[width=\columnwidth]{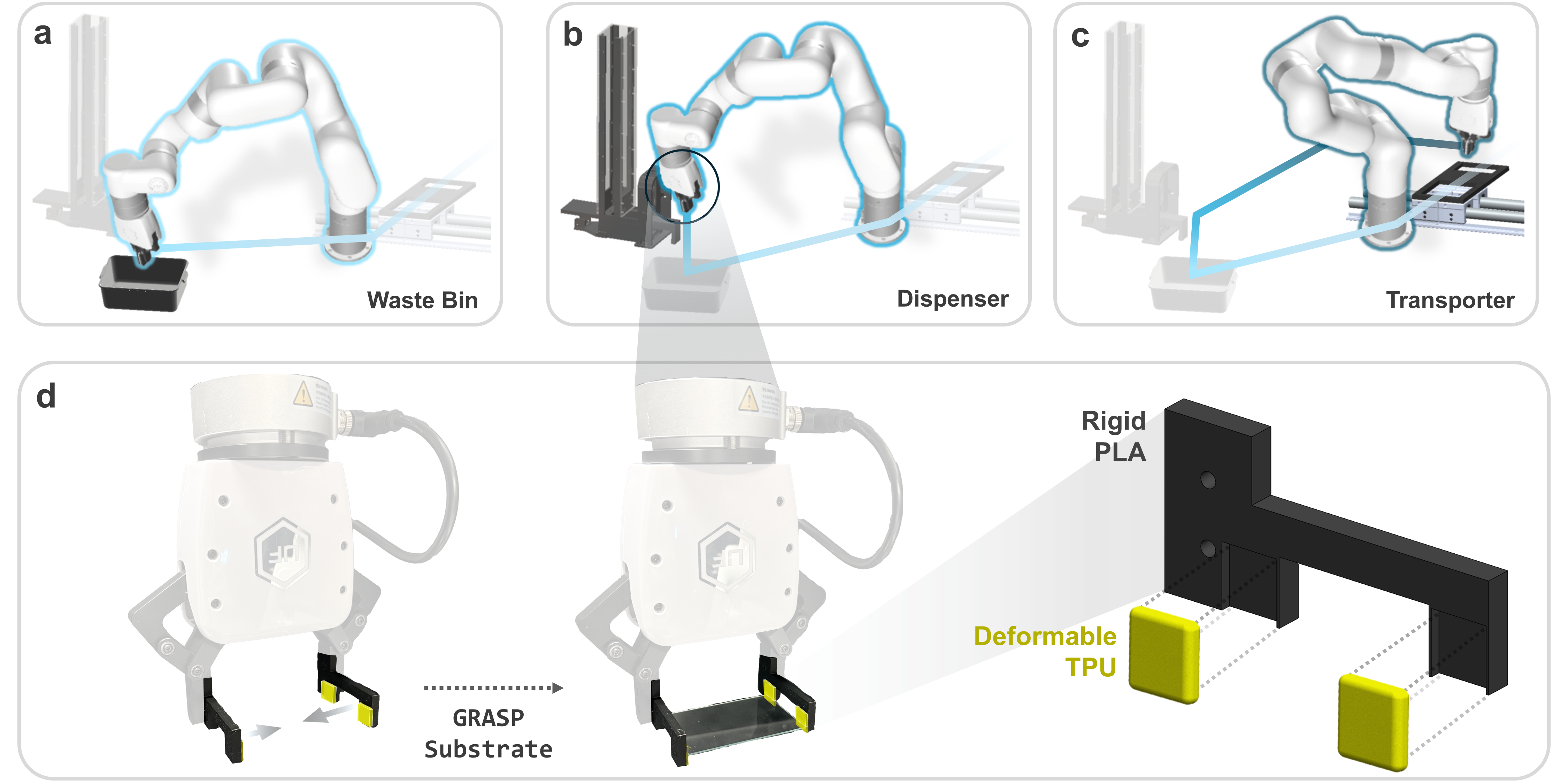 }
\end{subfigure}\hfill%
\caption{\footnotesize Robotic path plan for substrate reloading and grasping using specialized gripper fingers. The robotic arm path plan first (a) removes used substrates from the SDL transporter and moves them to a waste bin, then (b) picks up a fresh substrate from the dispenser, and finally (c) places the fresh substrate onto the SDL transporter. (d) Fragile substrates are manipulated with care, avoiding fracture through the use of a dual-material finger design that employs both rigid polylactic acid (PLA) and deformable thermoplastic polyurethane (TPU). Four points of contact are made between the substrate and the deformable TPU to ensure a secure grasp during robot motion.}
\label{fig:path}
\end{figure*}

With the advancement of computer vision methods and machine learning-based image segmentation methods, such as You Only Look Once (YOLO) \cite{redmon2016lookonceunifiedrealtime}, Segment Anything Model (SAM) \cite{kirillov2023segment}, and 3DSAM \cite{sam3dteam2025sam3d3dfyimages}, object detection has greatly improved in both accuracy and speed. However, transparent object detection and segmentation within images and environments remains a key challenge for these models due to the difficulty in resolving the edges of these objects in ambient lighting and obscuration from reflections, background interference, and refraction-induced distortions \cite{lai2015transparent, han2023segment, xie2020segmenting}. There have been numerous approaches to improve the detection of transparent objects, which include both traditional computer vision methods and deep learning methods \cite{khaing2019transparent, ruoning2024transparent, zheng2022glassnet}. Wang \textit{et al.} \cite{wang2024novel} develop a segmentation network architecture to address the challenge of capturing the texture characteristics of glass objects in images. Using this architecture, the authors demonstrate an improved performance in standard segmentation metrics such as mean absolute error, intersection union, weighted F-measure, and balanced error rate when compared to advanced glass segmentation methods such as MirrorNet \cite{yan2021mirrornet}, SAM \cite{kirillov2023segment}, and Translab \cite{xie2020segmenting}. Guo-Hua \textit{et al.} \cite{guohua2019transparent} propose a method that uses a combination of RGB, depth, and infrared images to address the difficulty in detecting transparent objects in different environments. Using this method, the author demonstrated the same accuracy in recognizing transparent objects as well as a 130\% reduction in falsely recognizing non-transparent objects, compared to the methods developed by Lysenkov \textit{et al.} \cite{lysenkov2013recognition}. Although these approaches address certain limitations in object segmentation and false recognition of transparent objects, they do not directly translate to the detection of transparent substrates in SDLs, as these existing methods largely focus on 3D and curved glass objects, such as beakers and flasks, allowing the models to take advantage of warping and distortion for easier detection. Transparent substrates used for materials experiments are often thin and flat with no curvature for these models to hone in on. Although there have been a few studies that combine computer vision and deep learning methods to find a balance between accuracy and speed for general object detection \cite{mathai2022transparent, rani2022object, wang2022detection}, the literature is sparse on demonstrating generalizability to transparent object detection. Rani \textit{et al.} \cite{rani2022object} use traditional computer vision techniques to strip down an image before feeding its wireframe into a Fast Region-based Convolutional Network (Fast R-NN) model for object detection and classification purposes. Using this methodology, the authors demonstrated improved accuracy with faster computation, compared to existing techniques. Pulli \textit{et al.} \cite{pulli2025enhancing} apply pre-processing techniques to images passed through a convolutional neural network for improved detection of transparent objects only using RGB images. Promising results using this approach are demonstrated on a user-trained GDR-Net \cite{wang2021gdr} framework, but the approach fails to improve performance using publicly accessible and pre-trained high-performance object detection models, such as YOLOX \cite{ge2021yolox}, limiting its generalizability. From these existing methods in the literature, joint computer vision and machine learning models for object detection have emerged as optimal architectures for improving general object detection accuracy and speed. However, further refinements and developments in these models are still required to reliably translate the methods to the detection of nearly featureless thin and flat transparent substrates for use within SDLs.

% , yet would need to be fine-tuned for accuracy for transparent substrates. Following the trend of increasing laboratory automation and usage of SDLs for materials experimentation and discovery \cite{tom2024selfdriving}, a need for intelligent and automated hardware-software systems that can image, segment, and handle delicate and transparent substrates has emerged.

Responding to this research gap, we propose a robotic approach of Automated Substrate Handling and Exchange (ASHE) of transparent substrates within SDLs that combines computer vision and deep learning for reliable substrate manipulation and error correction. ASHE utilizes a coupled hardware-software control approach to accurately and repeatably detect the placement of transparent substrates, precisely and carefully manipulate them, and then determine and correct any placement failures in a closed-loop, as shown in Figure \ref{fig:ashe}. On its hardware end, ASHE is equipped with a 5-degree-of-freedom (DOF) robotic arm, a specialized deformable gripper, and a dual-actuated dispenser to handle thin and fragile substrates. While on its software end, ASHE utilizes a fused geometric and deep learning model to reliably classify failed or successful transparent substrate placements by the robotic arm using lateral illumination techniques. With ASHE, we demonstrate a 98.5\% first-time success rate in the fully automated pick-and-place of fragile, transparent substrates across 130 experimental reload cycles of an SDL for materials discovery.

\section{Methods}
ASHE consists of three primary components: a robotic arm and gripper, an actuated substrate dispenser, and a deep learning error detection model. Each of these components interacts synchronously to remove the used transparent substrates from a holder within an SDL that transports the substrate between subsystems (here, we call it a \textit{transporter}), followed by dispensing and placing a new substrate on that transporter. Once placement has been completed, ASHE confirms if the new substrate has been placed successfully, and otherwise performs corrective re-positioning. Through a centralized integration of mechanical, electrical, and software systems, ASHE is commanded \textit{via} updates to a Structured Query Language (SQL) database, enabling modular integration into the existing architecture of an SDL wirelessly or through wired serial communication.

\subsection{Robotic Path Planning and Manipulation}

\begin{figure*}[ht!]
\centering
\begin{subfigure}{2\columnwidth}  
\includegraphics[width=\columnwidth]{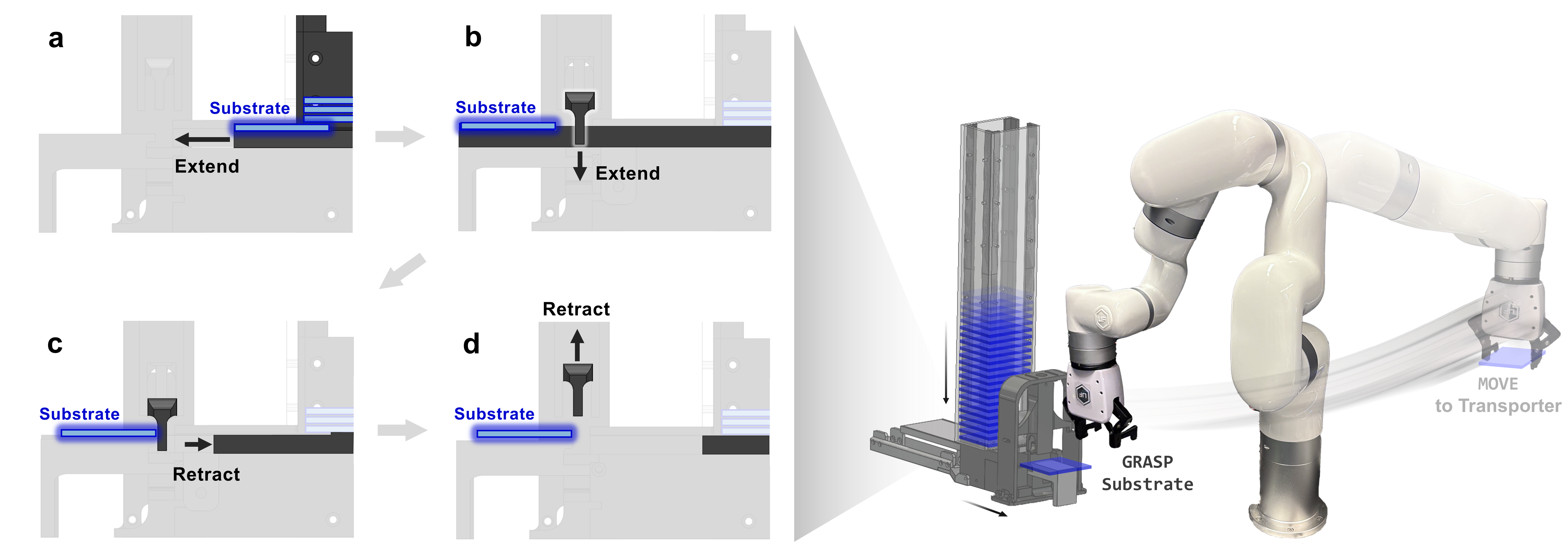 }
\end{subfigure}\hfill%
\caption{\footnotesize Dual-actuated dispenser sets fresh substrates for the robotic arm to grasp and move. (a) A single substrate is pushed from the bottom of a stack using horizontal actuation. (b) At the horizontal actuator's full extension, a vertical actuator is released to block the substrate from reentry. (c) The substrate stays in place as the horizontal actuator retracts. (d) The vertical actuator retracts, leaving the substrate in place for the robotic arm to grasp.}
\label{fig:dispense}
\end{figure*}

ASHE uses the UFACTORY xArm 5, a 5-DOF robotic arm, chosen for its Python compatibility, cost-effectiveness, and positional repeatability of 0.1 mm, which is important for the reliable grasping of fragile substrates. Programmed path plans of the robotic arm and subsequent substrate manipulations are activated using a leader-follower control approach, as dictated by updates to a binary SQL database. Hence, ASHE remains idle while the SDL state variable in the SQL database is set to a 0. To initiate ASHE, the transporter of the SDL moves to its target position and depresses a limit switch, sending a signal to the SQL database to change the SDL state variable to a 1, which then activates ASHE's path plan and concurrently dispenses a fresh substrate from the dual-actuated dispenser. First, the arm removes any used substrates from the SDL transporter and disposes them into a hazardous waste bin, as shown in Figure \ref{fig:path}a. Second, the arm moves to the dispenser to grasp a fresh substrate, as shown in Figure \ref{fig:path}b. Third, the arm moves to the transporter to place the fresh substrate, as shown in Figure \ref{fig:path}c, where computer vision is then used to determine if the placement was successful. 

 % A vertical actuator lifts the substrate out from the transporter to provide clearance for the robotic arm to grasp the used substrate for disposal.

To perform safe and successful grasping of fragile and brittle substrates such as glass and silica, a UFACTORY xARM gripper with custom-designed fingers is used for millimeter-scale control over the stroke and speed of the gripper actuation. Vacuum grippers were avoided since they make contact with the top of the substrates rather than the sides. Contacting the top of the substrate results in potential damage to the vacuum seal over time from the materials and chemicals deposited onto them by the SDL as well as the vacuum gripper interfering with any coatings applied to the fresh substrates. These custom-designed fingers are 3D printed and employ a dual-material design, combining polylactic acid (PLA) and thermoplastic polyurethane (TPU), to overcome the difficulty of grasping fragile substrates, as shown in Figure \ref{fig:path}d. Rigid PLA is used to provide structural integrity to the length of the finger for positioning precision, while deformable TPU 95A is used at the finger tips, providing compressibility during grasping contact to avoid fracture. We note that fingers with four points of contact on the sides of the substrate improve grasping when using deformable TPU tips, ensuring a more stable and secure grasp.

% ASHE is activated when the transporter arrives at a position near it. After an exchange between SQL databases, the transporter hits a limit switch before moving into its set position within the ASHE space. A signal is sent through the transporter's SQL database that the substrate is ready to be picked up; ASHE's SQL database receives this signal and sets the robot arm in motion. As the arm moves to pick up the substrate, the dispensing and lifting actuators are also triggered from the limit switch after a short delay to account for the SQL database exchange. The dispenser outputs a new substrate while the old substrate is lifted for easier accessibility for the robot arm. The robot arm picks up the old substrate from the transporter and disposes of it in a separate container. 

% To control which transporter will call ASHE to replenish its glass substrate, Structured Query Language (SQL) databases are used to control the interaction between ASHE and the transporters. One database holds the state of the robot arm (0 for neutral position, 1 for front transporter path plan, 2 for back transporter path plan); another database holds the state of each transporter and if it is ready to receive a new substrate. This interaction via SQL databases ensures that ASHE is only utilizing one path plan at a time and that it does not prematurely begin a path plan with the transporters out of position. 

\subsection{Dual-actuated Substrate Dispenser}

The dual-actuated substrate dispenser allows for ASHE to operate during long self-driving experimental campaigns by dispensing fresh substrates from a tall stack that can hold over 300 glass substrates of size 50.8 mm $\times$ 76.2 mm $\times$ 1.0 mm, while maintaining certain substrate surface treatments such as hydrophobic Teflon coating. Using a dual actuator design, this structure dispenses fresh substrates automatically, adhering to the same leader-follower control scheme as the robotic arm. This subsystem is controlled by an Arduino Due microprocessor with L293D motor drivers to control the positions of both linear actuators that dispense the substrates from the stack.

Figure \ref{fig:dispense} illustrates a cross-sectional view of the dispensing procedure. A vertical hotel stores a tall stack of fresh substrates, enabling the subsystem to have a small footprint. Within this hotel, the lower-most substrates rest atop a 3D printed transfer plate attached to the horizontally positioned linear actuator. This transfer plate is designed with a small lip approximately 0.2 mm thinner than the thickness of each susbtrate, enabling it to push only the lower-most substrate out from the hotel during actuation, even with the weight of the substrates above, as shown in Figure \ref{fig:dispense}a. When the horizontal actuator reaches its desired positioned, it depresses a limit switch, sending an electrical signal to activate a vertical actuator downward, as shown in Figure \ref{fig:dispense}b. This vertical actuator extends a 3D printed wall into the return path of the transfer plate's retraction to block the substrate from transferring back into the hotel, as shown in Figure \ref{fig:dispense}c. Finally, once the horizontal actuator is retracted, then the vertical actuator retracts and the system is back in its initial state, ready for its next command, as shown in Figure \ref{fig:dispense}d. The singular substrate now remains positioned on a platform to be picked up by the robotic arm for placement into the transporter of an SDL.

\begin{figure}[ht!]
\centering
\begin{subfigure}{0.9\columnwidth}  
\includegraphics[width=\columnwidth]{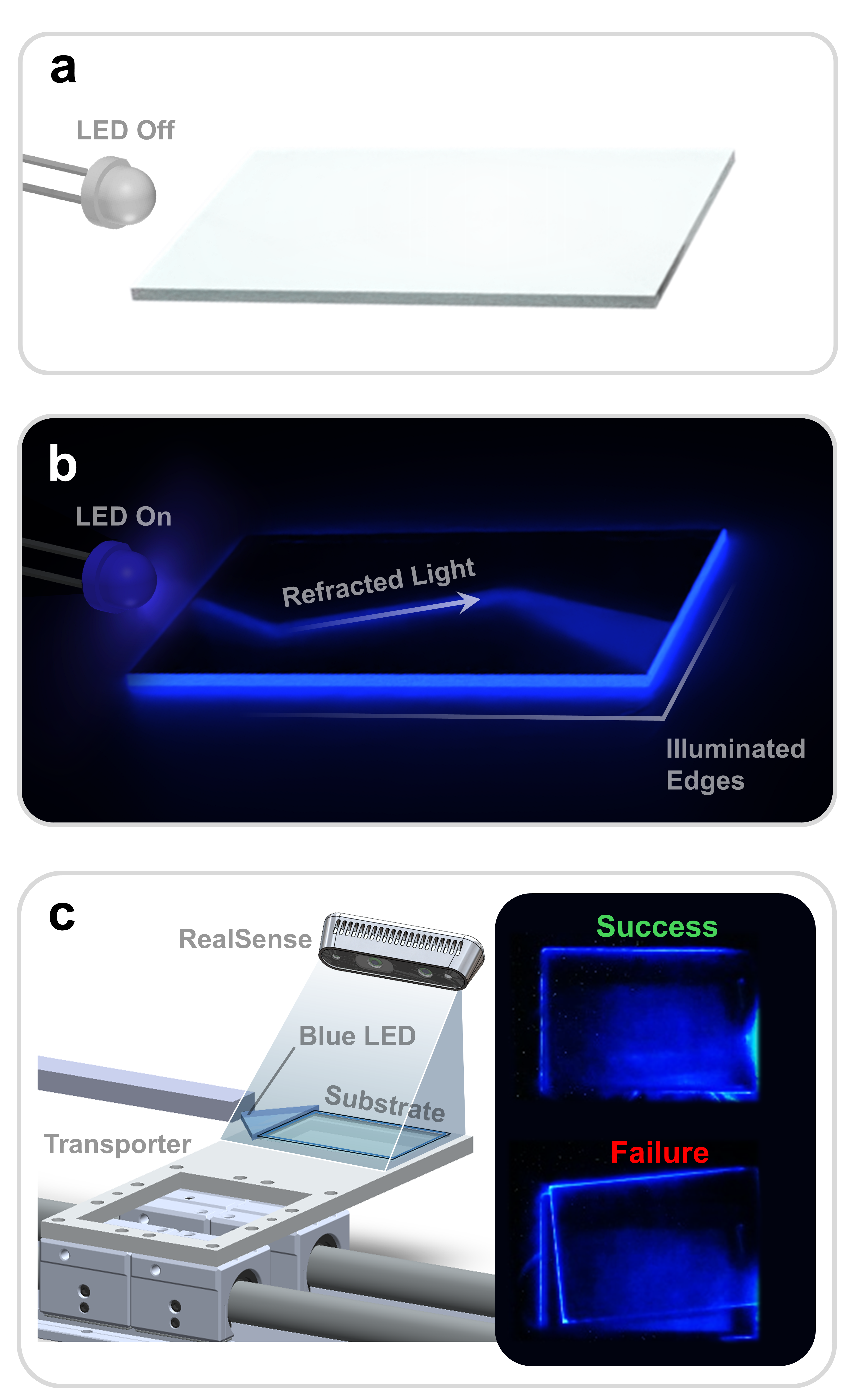 }
\end{subfigure}\hfill%
\caption{\footnotesize Lateral illumination of transparent substrates using blue light for computer vision detection. (a) Rendering of a transparent glass substrate in ambient light with no lateral illumination. (b) Rendering of a transparent glass substrate in the dark with blue light lateral illumination. The light refracts within the glass and emits along the edge faces of the substrate, illuminating them for clearer computer vision detection. (c) The computer vision imaging setup with RealSense D435 camera and lateral blue light illumination used for error detection in transparent substrate placement. Experimental images are shown of a transparent glass substrate successfully and unsuccessfully placed into the target slot of the transporter under the lateral illumination condition.}
\label{fig:led}
\end{figure}

\subsection{Transparent Substrate Detection}

% The camera takes an image of the empty transporter as the robot arm then picks up the new substrate and places it on the transporter. After the robot arm has moved away, the camera takes another image of transporter slot, this time with the substrate in view. The geometric model then determines if there is a successful placement from the alignment of the glass and aluminum estimated positions. Then, the machine learning model finds if there is a successful placement using its robust loop that takes in multiple images. If both models agree that there is a successful placement, ASHE continues on its normal workflow. But if at least one model calls a failure, ASHE enters a failure loop, in which the robot arm triggers the actuators on its own and repeats the process until a successful placement or it has been in this loop for 2 retries. If there hasn't been a successful placement after more than 2 retries, ASHE will hold position and wait for human intervention to confirm the placement. After getting a successful placement, ASHE’s SQL database sends a signal that it has completed its job and the experiment can continue. The transporter then moves back to where it came from.

Misplacing a substrate onto a target position of an SDL will result in a cascade of failures downstream, such as misaligned material dispensing or coating and erroneous characterization measurement readings. Thus, it is imperative that the substrates be placed correctly into this target position, such as the target slot of the transporter. In this paper, we focus on the detection of transparent substrates useful for conductive materials experimentation, as this gap has not yet been addressed in the literature. However, detection of transparent substrates is a challenging task due to the difficulty of resolving their edges in ambient lighting, as shown in Figure \ref{fig:led}a. Here, we circumvent many of these issues by using a method of laterally illuminating the transparent substrate from the side with blue light in the dark. Figure \ref{fig:led}b illustrates the use of lateral illumination to strongly illuminate a blue boundary around the edges of the substrate through the refraction and diffusion of blue light within the substrate. This blue edge improves computer vision detection of the substrate position, enabling more accurate and robust detection of placement errors, as shown in Figure \ref{fig:led}c. 

% The clear glass substrates are difficult to detect on camera with the overhead lights on due to glare on the surface and the lack of distinct edges when viewed through the camera. To circumvent this issue, the area was blacked out with only blue LEDs used to light up a focused portion of where the substrate placement occurred. These blue LEDs are attached to the dispenser circuit and turn on at the same time as the actuators move for timing purposes and limiting the usage of multiple disjoint electronic boards. With the LED placed at the same level as the transporter slot, the edges of the substrate are illuminated and show up well on camera. The glare issue is resolved from eliminating all other lighting sources from area.

\begin{figure}[h]
\centering
\begin{subfigure}{1\columnwidth}  
\includegraphics[width=\columnwidth]{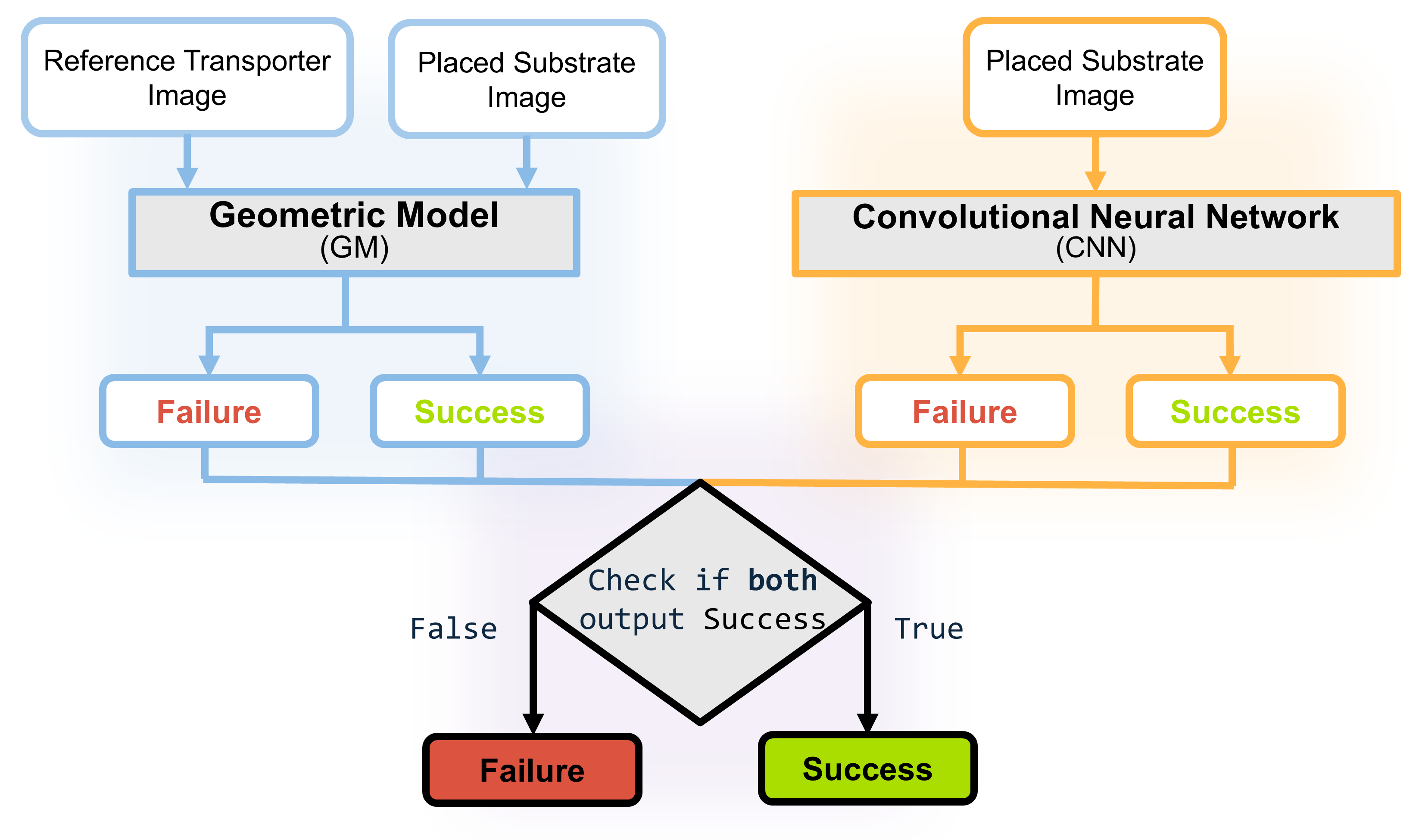 }
\end{subfigure}\hfill%
\caption{\footnotesize Fused geometric model (GM) and convolutional neural network (CNN) decision tree for reliable classification of placement errors for transparent substrates. A successful placement is declared only if the GM and CNN both output \texttt{success} classifications of the substrate placement.}
\label{fig:tree}
\end{figure}

\begin{figure*}[ht!]
\centering
\begin{subfigure}{2\columnwidth}  
\includegraphics[width=\columnwidth]{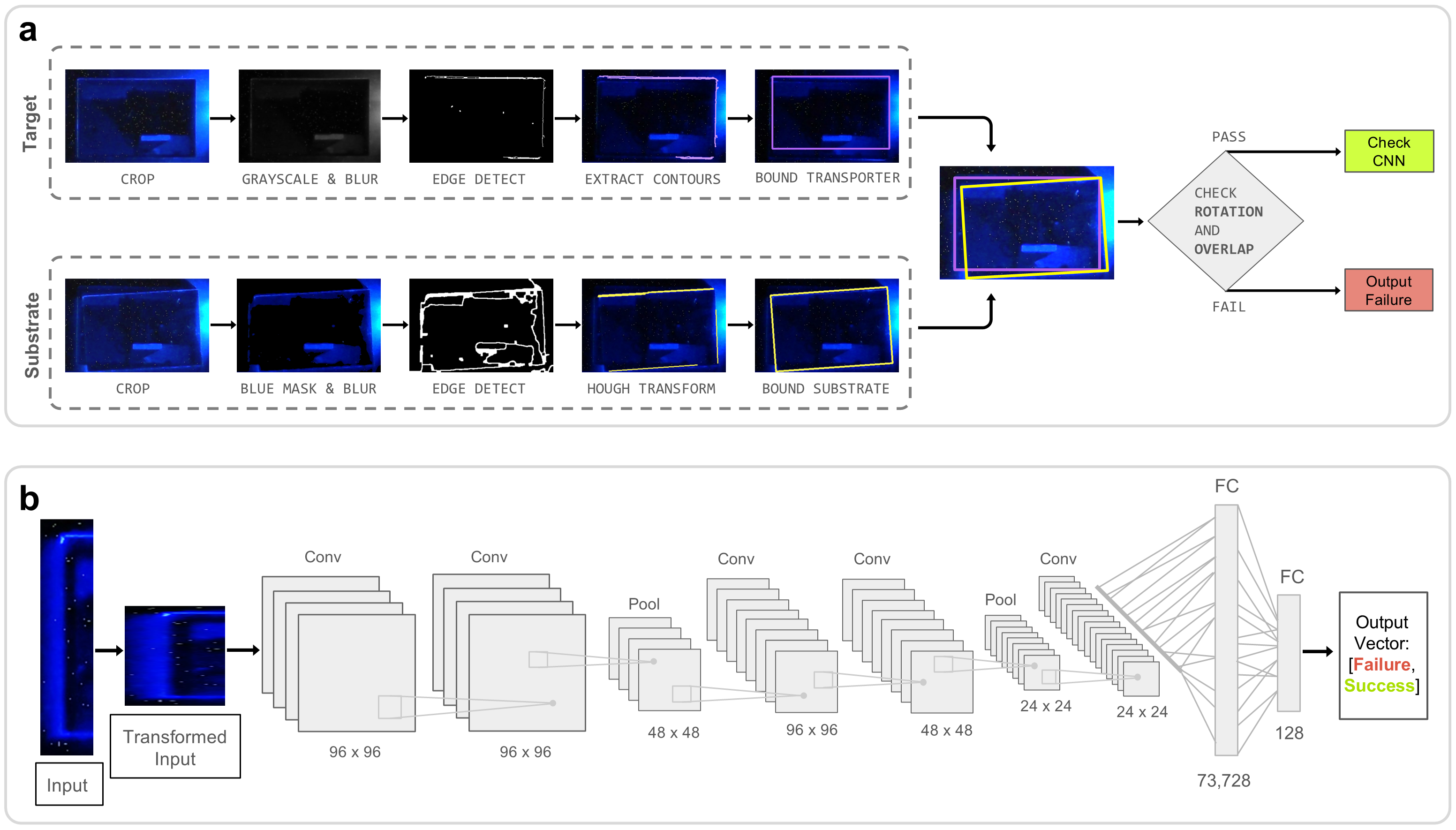 }
\end{subfigure}\hfill%
\caption{\footnotesize Workflow diagrams for the (a) geometric model (GM) and the (b) convolutional neural network (CNN) used for the detection of placement errors of transparent substrates. (a) The GM compares the bound edges created from both the placed substrate under lateral illumination and the target slot of the SDL transporter, where the substrate should be placed. Bound edges are formed through a suite of transformations, including color masking, edge detection, Hough transform, and convex hull transforms. If the reference and substrate bounding areas achieve an overlap greater than 90\%, then the GM outputs a \texttt{success} and checks the CNN for a corresponding \texttt{success}. (b) The CNN uses only one full edge and two partial edges of a laterally illuminated substrate image to classify the success of a placement. An image input to the CNN is reshaped into size 96 $\times$ 96 and passed through two 2D convolutional layers using 32 filters with 3 $\times$ 3 kernels. After max pooling, the feature map is passed through two more 2D convolutional layers using 64 filters with 3 $\times$ 3 kernels each, followed by a second max pooling. The fifth and final convolution uses 128 filters with a 3 $\times$ 3 kernel. The resulting feature map is flattened to a 1D vector and processed by a classification head with two fully connected layers and a 50\% dropout layer after the first fully connected layer to prevent overfitting. The CNN outputs a vector of length 2, where each output corresponds to the confidence of either a \texttt{failure} or a \texttt{success} classification, respectively $\in [0,1]$.}
\label{fig:pipe}
\end{figure*}

ASHE uses an Intel RealSense D435 camera to image the blue light-illuminated transparent substrates as placed into the transporter by the robotic arm. To detect whether the substrate has been successfully placed into the transporter, we propose a fused computer vision model approach. The first model uses the segmented geometry of the substrate compared to the target slot of the transporter to determine larger macro-scale errors in placement. The second model uses a deep learning convolutional neural network to determine smaller micro-scale errors in placement. Figure \ref{fig:tree} shows the decision tree of the fused model approach that enables robust and high-accuracy failure detection of transparent substrate placements. If either model detects a failed placement, then the robotic arm executes its auto-correction plan of removing the misplaced substrate and placing a freshly dispensed one, as previously shown in Figure \ref{fig:path}a--c.

\subsubsection{Geometric Detection of Macro-scale Errors}

The first part of the fused placement detection model for transparent substrates is a geometry-based computer vision model. This geometric model (GM) detects large placement errors by comparing the computed edges of both the target slot in the transporter and the transparent substrate. Figure \ref{fig:pipe}a illustrates this placement detection procedure of the GM.

First, the edges of the target slot in the transporter are detected. An image of the initial empty transporter is captured from the live feed of the aerial RealSense D435 before substrate placement. This image is then gray-scaled and blurred with a Gaussian filter to suppress noise in the surface texture. Then, Canny edge detection with a wide hysteresis threshold (30, 300) is performed \cite{canny2009computational}. The large upper threshold isolates strong structural boundaries, while the small lower threshold keeps relevant faint edges to account for uneven lighting conditions and shadows introduced by the blue light. Noisy edges are pruned if they do not meet a length threshold of at least 100 pixels. Finally, disconnected edges are closed by enforcing the known rigid geometry of the transporter using a convex hull transformation \cite{bradski2000opencv}, fitting the detected edges to a bounding rectangle.

Second, the edges of the placed transparent substrate are detected. This vision system takes advantage of the high-contrast lateral blue light illumination diffusely emitted from the edge faces of the substrate to perform edge detection. After robotic placement onto the transporter, a picture of the substrate is captured from the same reference frame as the initial empty transporter image using the aerial RealSense D435 camera. This image is blurred and filtered using a blue color mask to enhance the contrast around the edges of the substrate, which is then followed by Canny edge detection \cite{canny2009computational}. Using Canny edge detection on the color-masked image generates fragmented edges, resulting in inconsistent bounding rectangle results. Thus, a Hough transform \cite{ballard1981generalizing} is used to compute straight lines from the edge image. Finally, similar to determining the edges of the target slot in the transporter, the substrate edge segments are aggregated, approximated into a polygon, and bound to a convex hull \cite{bradski2000opencv} to construct the final bounding rectangle of the transparent substrate.

This segmentation process is repeated across 100 sampled frames taken from the RealSense D435 camera for both the target transporter slot and the transparent substrate to account for noise in the image and the scene in classifying placement success. Using the known approximate pixel dimensions of both the slot and substrate, only appropriately sized bounds are kept across these 100 computations. The average bounds of both the slot and the substrate are then computed and compared using Shapely \cite{shapely} to calculate the overlap percentage and relative rotation mismatch. A failed placement is output by the GM if shapes have either an area overlap of less than 90\%, determined through empirical testing. If the GM outputs a \texttt{failure} placement classification, then the robotic arm executes its auto-correction plan of removing the misplaced substrate and placing a freshly dispensed one, as previously shown in Figure \ref{fig:path}a--c.

\subsubsection{Deep Learning Detection of Micro-scale Errors}

% \begin{figure*}[ht!]
% \centering
% \begin{subfigure}{2\columnwidth}  
% \includegraphics[width=\columnwidth]{figs/cnn_architecture_R4.png}
% \end{subfigure}\hfill%
% \caption{Neural network architecture.}
% \label{fig:cnn}
% \end{figure*}

The second part of the fused placement detection model for transparent substrates is a 2D convolutional neural network (CNN). This CNN detects small placement errors through training on a labeled dataset of various rotational and translational placement errors. The collected dataset consists of 1990 captured images (996 unique failure cases and 994 success cases), prior to augmentation. For accurate micro-error detection, the raw RGB color is preserved for model training, and the images are cropped from 1920 $\times$ 1080 pixels into a fixed region of interest of 380 $\times$ 250 pixels along the far edge of the substrate from the blue LED. We note this cropping procedure improves CNN model performance by zooming in on regions containing more dense information on substrate placement accuracy and removes dead zones in the center of the substrate that provide little information on substrate position. Data augmentation was applied to improve model robustness to minor environmental changes in the input and to ensure the model could be generalized. These data augmentations included random shifts ($\pm$5 pixels), rotations ($\pm$1 degree), zooms ($\pm$2\%), brightness adjustment ($\pm$10), contrast adjustment ($\pm$5\%), and Gaussian noise ($\pm$5), generating 15 mutations per image, resulting in a total dataset size of 29,850 images. These augmentation values were selected to mimic realistic mechanical tolerances and sensor noise without changing key characteristics of the imaged substrate placement error. The dataset is split into 80\% training and 20\% validation subsets before augmentation to prevent data leakage. A weighted cross-entropy loss function was implemented to account for the data imbalance between success images and augmented failure images, reducing the rate of false positives by increasing attention to the minority class during model training. The model was trained using a learning rate of $1 \times 10^{-3}$ with a batch size of 32 over 50 epochs. Early stopping was implemented to prevent overfitting and to produce final model weights with the highest validation accuracy.

A shallow CNN architecture was created in PyTorch \cite{paszke2019pytorch} to learn micro-scale placement errors of the transparent substrates within the transporter under lateral illumination of blue light. Figure \ref{fig:pipe}b illustrates the micro-error detection CNN architecture designed for this study. The CNN architecture was designed to preserve spatial distribution during the early extraction of features, enabling the model to capture subtle spatial dependencies using the early feature maps. The network accepts a three-channel input tensor of 96 $\times$ 96 pixels (reshaped from the cropped 380 $\times$ 250 raw RGB image) and outputs a vector of length 2 from a classification head, where each output represents the classification confidence of a failure and success, respectively. During test time, ASHE utilizes the model to classify placement success with a confidence score across 100 uniquely sampled frames from the RealSense D435 camera live stream, returning the median placement success score. The placement success score is derived from the CNN's prediction confidence and is fitted onto a likelihood scale ranging from 0 (failure) to 1 (success). If the median achieves a placement success score of over 60\%, then the CNN part of the fusion model classifies the placement as a \texttt{success}. The placement success score threshold of 60\% was chosen after empirically testing various minuscule rotational and translation errors in substrate placement. If the CNN classifies the placement as a \texttt{failure}, then ASHE prompts the robotic arm to execute its auto-correction plan of removing the misplaced substrate and placing a freshly dispensed one, as previously shown in Figure \ref{fig:path}a--c.

% Therefore, the CNN model takes in an image of the placed substrate by the robotic arm and returns a success or failure classification label with corresponding confidence of the prediction. 

% The input tensor goes through two convolutions using 32 filters with 3 $\times$ 3 kernels and padding to maintain input dimensions. After two layers, the feature maps are downsized by a factor of two via max-pooling. The third and fourth convolution layers use 64 filters with 3 $\times$ 3 kernels each, followed by a second max-pooling layer which again reduces the map by a factor of two. The fifth and final convolution uses 128 filters with a 3 $\times$ 3 kernel.  The resulting feature map is flattened to a 1D vector and processed by a classification head with two fully connected layers. A 50\% dropout rate was applied after the first fully connected layer to prevent overfitting. 

\begin{figure*}[ht!]
\centering
\begin{subfigure}{2\columnwidth}  
\includegraphics[width=\columnwidth]{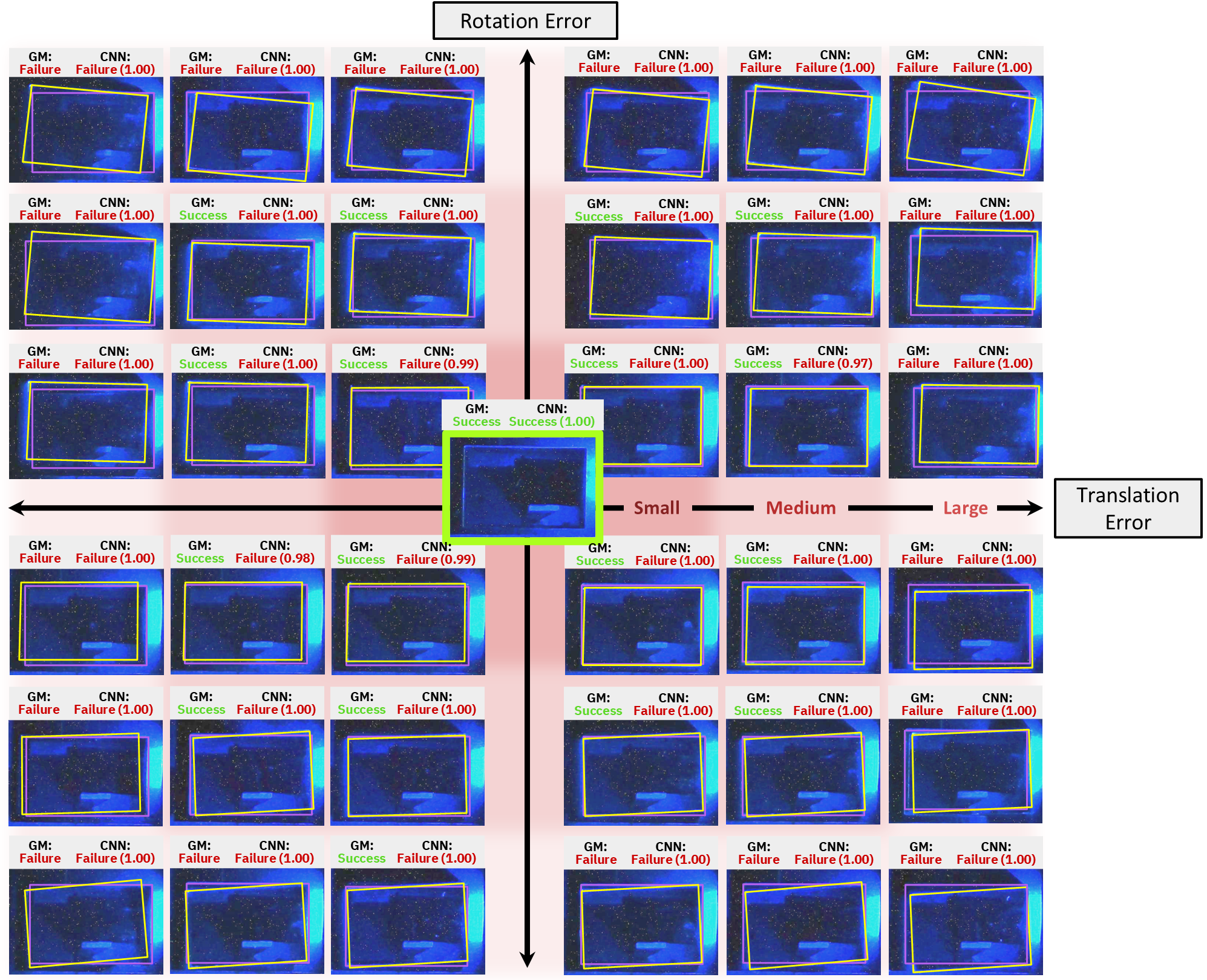 }
\end{subfigure}\hfill%
\caption{\footnotesize Transparent substrate placement error detection results of the geometric model (GM) and convolutional neural network (CNN) across an array of 36 unique placement error types and magnitudes. Each image within the matrix represents a unique placement error of a transparent substrate illuminated laterally by blue light. The yellow rectangle within each image represents the ground truth position of the substrate, while the purple rectangle represents the ground truth position of the SDL transporter target slot. Along the $X$-axis of the matrix, translational errors in placement increase in magnitude moving away from the center origin. Along the $Y$-axis of the matrix, rotational errors in placement increase in magnitude moving away from the center origin. The outermost ring of 20 images in the matrix represents placement errors with large translational and/or rotational errors. The middle ring of 12 images in the matrix represents placement errors with medium translational and/or rotational errors. The inner ring of 4 images in the matrix represents placement errors with small translational and/or rotational errors. The innermost image represents a successful placement. The output GM and CNN classifications are indicated above each image, along with the CNN classification confidence for that image. The yellow and purple bounding rectangles are for visualization purposes only and are not present in the raw data passed to the models for classification.}
\label{fig:fuse}
\end{figure*}

\section{Results}

\subsection{Transparent Substrate Vision Detection Performance}

Using the fused GM-CNN model, we demonstrate that reliable detection of transparent substrate placement failures is achieved across an array of realistic error modes. The GM is used as the initial rejection model for large, obvious errors, while the CNN is used to determine small errors nearly imperceptible to the human eye. However, a successful placement is not confirmed unless the GM and CNN both classify the placement as a \texttt{success}. 

Figure \ref{fig:fuse} illustrates both the GM and CNN model performance across 36 unique transparent glass substrate placement errors, including small, medium, and large errors in both translation and rotation. From these results, we highlight that the GM and CNN perform similarly on both rotational errors and translational errors. However, dramatic performance differences between the models can be seen across the small, medium, and large errors. For rotation, \textit{small errors} in substrate placement have an angle mismatch with the target transporter slot of less than approximately 1.7$^\circ$, while \textit{large errors} have an angle mismatch greater than approximately 5.4$^\circ$. For both horizontal and vertical translation, \textit{small errors} in substrate placement have a displacement less than approximately 2.4 mm, while \textit{large errors} have a displacement greater than approximately 5.6 mm.

In Figure \ref{fig:fuse}, the placement success classifications for both the GM and CNN models are shown above every image. The GM correctly detects $\frac{19}{20}$ of the large errors with only one false positive that has a counter-clockwise rotational error of 1.8$^\circ$ and a translational error of 2.7 mm. The CNN correctly detects all $\frac{20}{20}$ of the large errors with a prediction confidence of 100\% for all 20 large error images. Moving into the medium errors, the GM correctly detects $\frac{0}{12}$ of the medium errors, outputting false positives for all 12 images. The CNN correctly detects all $\frac{12}{12}$ of the medium errors with a prediction confidence ranging from 97\% to 100\%. Lastly, for the small errors, the GM correctly detects $\frac{0}{4}$ of the small errors, outputting false positives for all 4 images. The CNN correctly detects all $\frac{4}{4}$ of the small errors with a prediction confidence ranging from 99\% to 100\%. Both the GM and CNN correctly classify a \texttt{success} for the successful placement image.

\begin{figure*}[ht!]
\centering
\begin{subfigure}{1.6\columnwidth}  
\includegraphics[width=\columnwidth]{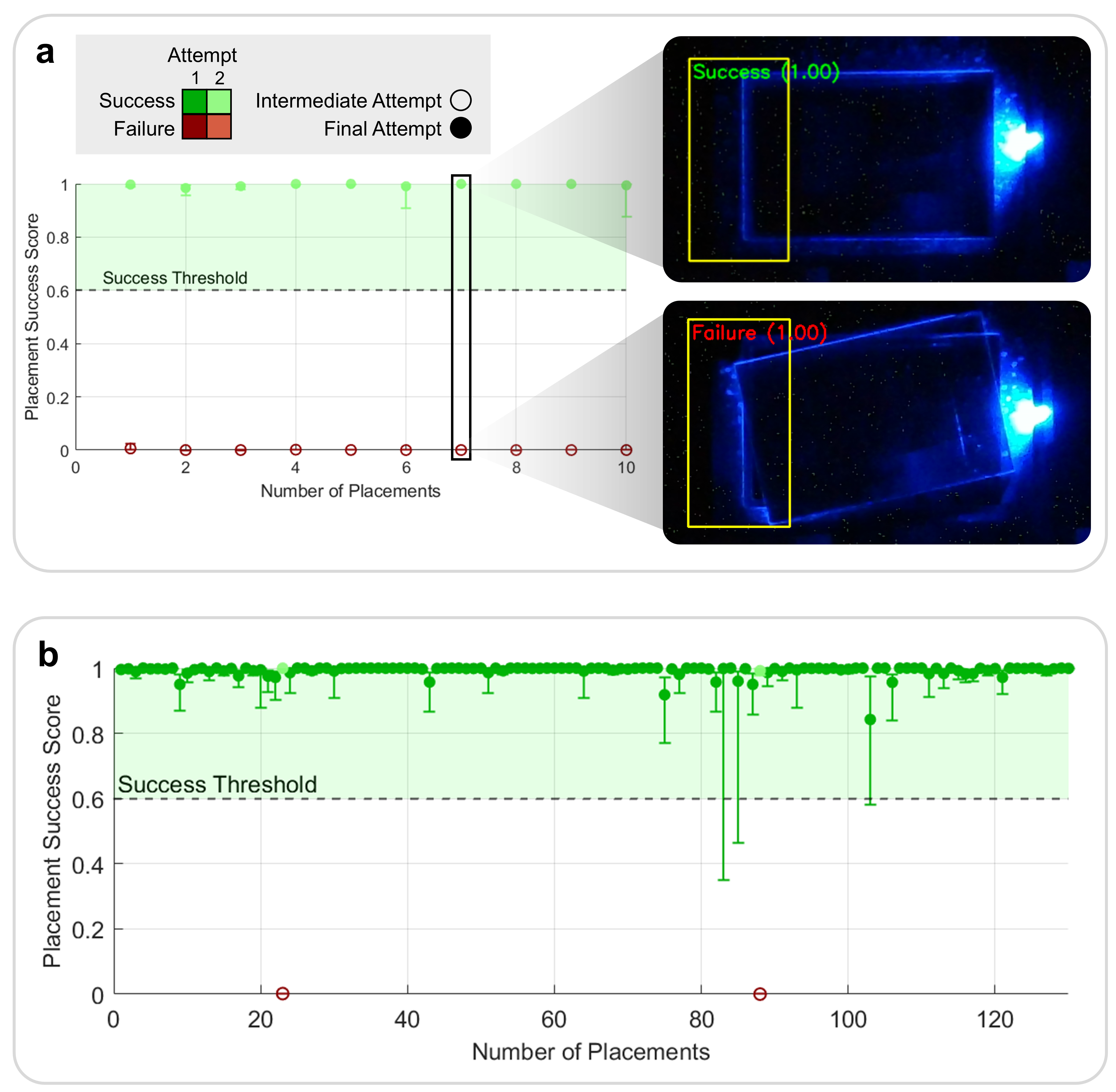}
\end{subfigure}\hfill%
\caption{\footnotesize Experimental results of the closed-loop ASHE system reloading transparent glass substrates within an SDL. (a) Computer vision error detection and correction performance of ASHE across 10 pre-programmed failures in robotic arm placement of transparent substrates. (b) Long-form repeatability tests across 130 independent cycles of transparent glass substrate pick-and-placement using ASHE, without pre-programmed failures. Each scatter point represents a placement success score of a freshly dispensed substrate into the target slot of an SDL transporter by the robotic arm. The placement success score is derived from the CNN's prediction confidence and is fitted onto a likelihood scale ranging from 0 (failure) to 1 (success). An empty scatter point represents an intermediate attempt at placement, while a full scatter point represents the final attempt at placement. The shade of these scatter points represents the placement attempt: a darker shade for the first placement attempt and a lighter shade for the second placement attempt. The color of the scatter point represents the median confidence score of the fused GM-CNN model, with error bars illustrating the 25th and 75th percentiles across 100 predictions, each on a unique frame taken from the RealSense D435 camera live stream after placement. For every trial, the vision classification is validated by a human domain expert to correctly match the real-world failure or successful placement.}
\label{fig:results}
\end{figure*}

\subsection{ASHE Repeatability Performance}

Combining the robotic arm with specialized deformable grippers, the dual-actuated substrate dispenser, and the fused GM-CNN vision detection model, we experimentally validate the performance of the complete closed-loop ASHE pipeline. In these experiments, each trial represents an independent reloading procedure performed by ASHE, activated through the leader-follower control approach when the SDL transporter reaches its reloading position and depresses a limit switch, changing its SQL state variable from 0 to 1, in turn commanding ASHE to begin the reloading procedure. Figure \ref{fig:results} shows the results of these repeatability experiments running on the complete ASHE system.

Figure \ref{fig:results}a highlights the closed-loop failure detection and auto-correction capabilities of ASHE on 10 pre-programmed robotic arm substrate placement failures. The dark red open scatter points indicate ASHE's first attempt at placement into the SDL transporter, which are pre-programmed to fail, and, here, ASHE correctly classifies $\frac{10}{10}$ as \texttt{failure}. The light green solid scatter points indicate ASHE's second placement attempt to correct these detected failures of the first attempt. We demonstrate that $\frac{10}{10}$ of the automated corrective procedures successfully corrected the placement failures on the second placement attempt, resulting in \texttt{success} classifications by the vision system. 

% Moreover, the dark green solid scatter points indicate the standard operational performance of ASHE to correctly dispense and place transparent glass substrates into the targeted SDL transporter slot. We demonstrate that $\frac{40}{40}$ of these placed substrates are correctly detected as successful placements by the vision system.

Figure \ref{fig:results}b demonstrates the long-form automated reliability and repeatability performance of ASHE across 130 uniquely commanded reloading procedures, without any pre-programmed failures. ASHE classifies \texttt{success} placements on $\frac{128}{130}$ of the reload cycles on the first attempt. For every placement, a human domain expert monitors the placement result, and all $\frac{128}{130}$ are validated as correctly classified. Therefore, across these 130 experiments, ASHE achieves a 98.5\% first-time placement accuracy. ASHE classifies the remaining $\frac{2}{130}$ substrates as \texttt{failure}, which are validated by the human domain expert. Upon this \texttt{failure} classification, ASHE automatically corrects the failures and places fresh substrates, classifying these second attempts as \texttt{success}, which are validated by the human domain expert. Ultimately, through the closed-loop detection and auto-correction capabilities of ASHE, 100\% of the substrates are placed correctly into the target slot of the SDL transporter by the second attempt. These results are noted to transfer to transparent substrates with various non-streaking and transparent surface treatments, such as hydrophobic Teflon coatings and hydrophilic plasma treatments.

\section{Discussion and Conclusions}

In this current expansion of automated methods for materials discovery and self-driving laboratories \cite{tom2024selfdriving, macleod2020selfdriving, crabtree2020selfdriving}, improving system accuracy and repeatability is important to further the advancement and success of these laboratories. With the foundation of materials and chemical sciences starting with a substrate to transfer or dispense materials onto \cite{venables2000introduction}, it is critical that the infrastructure of these self-driving laboratories be developed to handle and manipulate a wide variety of substrates with high accuracy and repeatability. The most challenging substrates for automated methods to manipulate are fragile, thin, and transparent substrates, such as glass or silica, used for conductive materials \cite{fraser1972highly, bash2021multi, ma2019graphene, um2025tailoring, siemenn2024using, sun2021data}, since current methods fail to detect or sufficiently generalize to these nearly featureless substrates \cite{khaing2019transparent, ruoning2024transparent, zheng2022glassnet}. Thus, a research gap exists in designing software capabilities to accurately detect these transparent substrates as well as hardware capabilities to reliably manipulate them, while automatically correcting for failed placements.

% Furthermore, the foundation of materials science and chemistry experiments begins with a substrate to deposit materials onto \cite{venables2000introduction}. Common substrates include transparent silica or glass, often used to deposit electrically conductive materials onto . Historically, transparent objects, such as these silica or glass substrates, have been immensely challenging to image and detect accurately using computer vision and machine learning tools . Thus, a research gap exists in reliably and robustly automating the manipulation and reloading of transparent substrates for self-driving materials laboratories. 

In this paper, we present a fully closed-loop system for automating transparent substrate handling and exchange in self-driving laboratories, entitled ASHE. ASHE synchronously integrates a 5-degree-of-freedom robotic arm with a specialized deformable gripper, a dual-actuated substrate dispenser, and a fused geometric and deep learning vision detection model to accurately unload used substrates and load fresh substrates with full automation within a self-driving laboratory. The fused geometric and deep learning vision system in ASHE enables micro-scale error detection and closed-loop correction of substrate misplacements. In this paper, we show that the fused model approach accurately classifies small, medium, and large translational and rotational placement errors of transparent substrates with classification confidence ranging from 97\% to 100\%. We run 130 independent trials of ASHE reloading substrates into a self-driving laboratory and demonstrate 98.5\% first-time placement accuracy, validated by a human domain expert, with only two substrate misplacements. ASHE automatically detects these misplaced substrates and automatically corrects their placements.

% Despite these repeatability tests revealing high placement success rates, incorporating vision detection software and a failure recovery system remains necessary for long, autonomous materials discovery campaigns. 

We demonstrate that ASHE is robust to false positives and tends to overgeneralize classifications of successfully placed substrates as \texttt{failure} instead. This behavior is advantageous as it mitigates the risk of passing a misplaced substrate through to the self-driving laboratory, which would likely result in a cascade of downstream failures if the substrate is not properly set into its target position. False negatives are preferable to encounter using ASHE compared to the alternative false positive, since the automated error correction mechanism is implemented to handle these instances by automatically activating reloading protocol upon a \texttt{failure} classification.

Although ASHE demonstrates promising performance results towards the advancement of automated transparent substrate manipulation, vision detection, and error correction, several limitations exist in its generalizability. Specifically, the hardware and software components of the system have been designed and trained to perform highly on glass substrates of a particular size of 50.8 mm $\times$ 76.2 mm $\times$ 1.0 mm. To extend the use of ASHE to substrates outside of these parameters, the actuated substrate dispenser and gripper fingers must be redesigned, and the deep learning model must be retrained on a newly collected dataset for that particular substrate. Furthermore, wider adoption of the technologies used to develop ASHE may be inhibited by their costs. Although many of the components of ASHE are 3D printed, the main cost sinks include the UFACTORY xARM 5 at \$6000 USD, and its corresponding gripper at \$2300 USD. While these costs are on the lower end of robotic arms, acquiring hardware with sufficient precision capabilities governs the overall success of the system.

Ultimately, ASHE helps to close the research gap in fragile and transparent substrate manipulation. Through the integration of robotic arms, automated substrate dispensing, and reliable vision detection, closed-loop control and error correction are established. With ASHE, we move towards an improvement in the automation capabilities of self-driving materials laboratories with accurate, reliable, and robust robotic manipulation for fragile and transparent substrates. 

% Because of the many different ways the placement of substrates can fail, the failure recovery path plan was developed to accommodate these wide ranges that could be expected. To account for rotational errors, the gripper rotates 90 degrees before picking up the misplaced substrate by the long side. As the gripper closes, the substrate is aligned from its rotation and firmly grasped. This method also allows for the grasp of large translational errors since the gripper can now reach more area.

% \section{Conclusion}

% In this paper, we highlight the design and development of ASHE, a downstream component of self-driving laboratories that use glass substrates for high-throughput experiments. The key design features of ASHE include a robotic arm with precise path plans, a custom dual material gripper for secure grasp of delicate substrates, an automated substrate dispenser, and a software pipeline that can detect different levels of errors in the substrate placement. 

\section*{Author contributions}
A.E.S. and T.B. conceptualized the work. K.F. and A.E.S. designed the methodology. K.F., I.G., and A.E.S. designed and built the hardware and electronics. K.F. and A.G. wrote the software. K.F., A.G., and I.G. conducted experiments. K.F., A.G., and A.E.S. wrote the manuscript. All authors reviewed and edited the manuscript. A.E.S. and T.B. provided guidance.

\section*{Conflicts of interest}
There are no conflicts to declare.

\section*{Data availability}

The code and data presented in this article are available on GitHub at \href{https://github.com/PV-Lab/ASHE}{https://github.com/PV-Lab/ASHE}.

\section*{Acknowledgements}

The authors acknowledge the support of the Massachusetts Institute of Technology Undergraduate Research Opportunities Program (UROP) office and the MIT Energy Initiative (MITEI) program. The authors acknowledge Eunice Aissi and Alexander Love for their fruitful discussions on system design modifications and improvements.

%%%END OF MAIN TEXT%%%


\balance

\renewcommand\refname{References}

%%%REFERENCES%%%
\bibliography{references} %You need to replace "rsc" on this line with the name of your .bib file

@article{um2025tailoring,
  title={Tailoring Molecular Space to Navigate Phase Complexity in Cs-Based Quasi-2D Perovskites via Gated-Gaussian-Driven High-Throughput Discovery},
  author={Um, Minsub and Sanchez, Sheryl L and Song, Hochan and Lawrie, Benjamin J and Ahn, Hyungju and Kalinin, Sergei V and Liu, Yongtao and Choi, Hyosung and Yang, Jonghee and Ahmadi, Mahshid},
  journal={Advanced Energy Materials},
  volume={15},
  number={16},
  pages={2404655},
  year={2025},
  publisher={Wiley Online Library}
}

@article{bash2021multi,
  title={Multi-fidelity high-throughput optimization of electrical conductivity in P3HT-CNT composites},
  author={Bash, Daniil and Cai, Yongqiang and Chellappan, Vijila and Wong, Swee Liang and Yang, Xu and Kumar, Pawan and Tan, Jin Da and Abutaha, Anas and Cheng, Jayce JW and Lim, Yee-Fun and others},
  journal={Advanced Functional Materials},
  volume={31},
  number={36},
  pages={2102606},
  year={2021},
  publisher={Wiley Online Library}
}

@article{lu2007constrained,
  title={Constrained sintering of YSZ/Al2O3 composite coatings on metal substrates produced from eletrophoretic deposition},
  author={Lu, X-J and Xiao, P},
  journal={Journal of the European Ceramic Society},
  volume={27},
  number={7},
  pages={2613--2621},
  year={2007},
  publisher={Elsevier}
}

@article{bai2019accelerated,
  title={Accelerated discovery of organic polymer photocatalysts for hydrogen evolution from water through the integration of experiment and theory},
  author={Bai, Yang and Wilbraham, Liam and Slater, Benjamin J and Zwijnenburg, Martijn A and Sprick, Reiner Sebastian and Cooper, Andrew I},
  journal={Journal of the American Chemical Society},
  volume={141},
  number={22},
  pages={9063--9071},
  year={2019},
  publisher={ACS Publications}
}

@article{matsuda2022data,
  title={Data-driven automated robotic experiments accelerate discovery of multi-component electrolyte for rechargeable Li--O2 batteries},
  author={Matsuda, Shoichi and Lambard, Guillaume and Sodeyama, Keitaro},
  journal={Cell Reports Physical Science},
  volume={3},
  number={4},
    pages = {100832},
  year={2022},
  publisher={Elsevier}
}

@article{sun2021data,
  title={A data fusion approach to optimize compositional stability of halide perovskites},
  author={Sun, Shijing and Tiihonen, Armi and Oviedo, Felipe and Liu, Zhe and Thapa, Janak and Zhao, Yicheng and Hartono, Noor Titan P and Goyal, Anuj and Heumueller, Thomas and Batali, Clio and others},
  journal={Matter},
  volume={4},
  number={4},
  pages={1305--1322},
  year={2021},
  publisher={Elsevier}
}

@article{siemenn2024using,
  title={Using scalable computer vision to automate high-throughput semiconductor characterization},
  author={Siemenn, Alexander E and Aissi, Eunice and Sheng, Fang and Tiihonen, Armi and Kavak, Hamide and Das, Basita and Buonassisi, Tonio},
  journal={Nature Communications},
  volume={15},
  number={1},
  pages={4654},
  year={2024},
  publisher={Nature Publishing Group UK London}
}

@article{ma2019graphene,
  title={Graphene-based transparent conductive films: material systems, preparation and applications},
  author={Ma, Yingjie and Zhi, Linjie},
  journal={Small Methods},
  volume={3},
  number={1},
  pages={1800199},
  year={2019},
  publisher={Wiley Online Library}
}

@article{fraser1972highly,
  title={Highly conductive, transparent films of sputtered In2- xSnx O 3- y},
  author={Fraser, DB and Cook, HD},
  journal={Journal of the Electrochemical Society},
  volume={119},
  number={10},
  pages={1368},
  year={1972},
  publisher={IOP Publishing}
}

@article{lee2023fully,
    author ={Lee, Jules and Mulay, Prajakatta and Tamasi, Matthew J. and Yeow, Jonathan and Stevens, Molly M. and Gormley, Adam J.},
    title  ={A fully automated platform for photoinitiated RAFT polymerization},
    journal  ={Digital Discovery},
    year  ={2023},
    volume  ={2},
    issue  ={1},
    pages  ={219-233},
    publisher  ={RSC}
}

@Article{jiang2023autonomous,
    author ="Jiang, Ying and Fakhruldeen, Hatem and Pizzuto, Gabriella and Longley, Louis and He, Ai and Dai, Tianwei and Clowes, Rob and Rankin, Nicola and Cooper, Andrew I.",
    title  ="Autonomous biomimetic solid dispensing using a dual-arm robotic manipulator",
    journal  ="Digital Discovery",
    year  ="2023",
    volume  ="2",
    issue  ="6",
    pages  ="1733-1744",
    publisher  ="RSC"
}

@Article{yotsumoto2024autonomous,
    author ="Yotsumoto, Yuto and Nakajima, Yusaku and Takamoto, Ryusei and Takeichi, Yasuo and Ono, Kanta",
    title  ="Autonomous robotic experimentation system for powder X-ray diffraction",
    journal  ="Digital Discovery",
    year  ="2024",
    volume  ="3",
    issue  ="12",
    pages  ="2523-2532",
    publisher  ="RSC",
}

@Article{fernando2025robotic,
    author ="Fernando, Chandima and Marcello, Hailey and Wlodek, Jakub and Sinsheimer, John and Olds, Daniel and Campbell, Stuart I. and Maffettone, Phillip M.",
    title  ="Robotic integration for end-stations at scientific user facilities",
    journal  ="Digital Discovery",
    year  ="2025",
    volume  ="4",
    issue  ="4",
    pages  ="1083-1091",
    publisher  ="RSC",
}

@Article{sadeghi2025selfdriving,
    author ="Sadeghi, Sina and Mattsson, Karl and Glasheen, Joshua and Lee, Victoria and Stark, Christine and Jha, Pragyan and Mukhin, Nikolai and Li, Junbin and Ghorai, Arup and Orouji, Negin and Moran, Christopher H. J. and Velayati, Alireza and Bennett, Jeffrey A. and Canty, Richard B. and Reyes, Kristofer G. and Abolhasani, Milad",
    title  ="A self-driving fluidic lab for data-driven synthesis of lead-free perovskite nanocrystals",
    journal  ="Digital Discovery",
    year  ="2025",
    volume  ="4",
    issue  ="7",
    pages  ="1722-1733",
    publisher  ="RSC",
}

@Article{sorkun2025redcat,
    author ="Sorkun, Murat Cihan and Zhou, Xuan and Murigneux, Joannes and Menegazzo, Nicola and Narsaria, Ayush Kumar and Thanoon, David and Klusener, Peter A. A. and Kaluskar, Kaustubh and Shetty, Sharan and Barmpoutsis, Efstathios and Er, Süleyman",
    title  ="RedCat{,} an automated discovery workflow for aqueous organic electrolytes",
    journal  ="Digital Discovery",
    year  ="2025",
    volume  ="4",
    issue  ="7",
    pages  ="1844-1855",
    publisher  ="RSC",
}

@Article{pickles2025automated,
    author ="Pickles, Thomas and Leghrib, Youcef and Weisshaar, Matt and Goncharuk, Mikhail and Timperman, Peter and Doherty, Timothy and Ford, David D. and Moores, Jonathan and Florence, Alastair J. and Brown, Cameron J.",
    title  ="Automated scale-up crystallisation DataFactory for model-based pharmaceutical process development: a Bayesian case study",
    journal  ="Digital Discovery",
    year  ="2025",
    volume  ="4",
    issue  ="8",
    pages  ="2025-2032",
    publisher  ="RSC",
}

@Article{yik2023automated,
    author ="Yik, Jackie T. and Zhang, Leiting and Sjölund, Jens and Hou, Xu and Svensson, Per H. and Edström, Kristina and Berg, Erik J.",
    title  ="Automated electrolyte formulation and coin cell assembly for high-throughput lithium-ion battery research",
    journal  ="Digital Discovery",
    year  ="2023",
    volume  ="2",
    issue  ="3",
    pages  ="799-808",
    publisher  ="RSC",
}

@Article{soh2023automated,
    author ="Soh, Beatrice W. and Chitre, Aniket and Lee, Wen Yang and Bash, Daniil and Kumar, Jatin N. and Hippalgaonkar, Kedar",
    title  ="Automated pipetting robot for proxy high-throughput viscometry of Newtonian fluids",
    journal  ="Digital Discovery",
    year  ="2023",
    volume  ="2",
    issue  ="2",
    pages  ="481-488",
    publisher  ="RSC",
}

@Article{nishio2025digital,
    author ="Nishio, Kazunori and Aiba, Akira and Takihara, Kei and Suzuki, Yota and Nakayama, Ryo and Kobayashi, Shigeru and Abe, Akira and Baba, Haruki and Katagiri, Shinichi and Omoto, Kazuki and Ito, Kazuki and Shimizu, Ryota and Hitosugi, Taro",
    title  ="A digital laboratory with a modular measurement system and standardized data format",
    journal  ="Digital Discovery",
    year  ="2025",
    volume  ="4",
    issue  ="7",
    pages  ="1734-1742",
    publisher  ="RSC",
}

@article{szymanski2023autonomous,
    author = {Szymanski, Nathan J. and Rendy, Bernardus and Fei, Yuxing and Kumar, Rishi E. and He, Tanjin and Milsted, David and McDermott, Matthew J. and Gallant, Max and Cubuk, Ekin Dogus and Merchant, Amil and Kim, Haegyeom and Jain, Anubhav and Bartel, Christopher J. and Persson, Kristin and Zeng, Yan and Ceder, Gerbrand},
	title = {An autonomous laboratory for the accelerated synthesis of novel materials},
    journal = {Nature},
    year = {2023},
	volume = {624},
	issn = {1476-4687},
	pages = {86--91},
}

@Article{knox2025selfdriving,
    author ="Knox, Stephen T. and Wu, Kai E. and Islam, Nazrul and O{'}Connell, Roisin and Pittaway, Peter M. and Chingono, Kudakwashe E. and Oyekan, John and Panoutsos, George and Chamberlain, Thomas W. and Bourne, Richard A. and Warren, Nicholas J.",
    title  ="Self-driving laboratory platform for many-objective self-optimisation of polymer nanoparticle synthesis with cloud-integrated machine learning and orthogonal online analytics",
    journal  ="Polym. Chem.",
    year  ="2025",
    volume  ="16",
    issue  ="12",
    pages  ="1355-1364",
    publisher  ="RSC",
}

@article{wang2025autonomous,
	title = {Autonomous platform for solution processing of electronic polymers},
    author = {Wang, Chengshi and Kim, Yeon-Ju and Vriza, Aikaterini and Batra, Rohit and Baskaran, Arun and Shan, Naisong and Li, Nan and Darancet, Pierre and Ward, Logan and Liu, Yuzi and Chan, Maria K. Y. and Sankaranarayanan, Subramanian K.R.S. and Fry, H. Christopher and Miller, C. Suzanne and Chan, Henry and Xu, Jie},
    journal = {Nature Communications},
	volume = {16},
    number = {1},
    pages = {1498},
	year = {2025},
}

@article{crabtree2020selfdriving,
    author = {Crabtree, George},
    title = {Self-Driving Laboratories Coming of Age},
    journal = {Joule},
    year = {2020},
    volume = {4},
    pages = {2538-2541}
}

@article{zhang2024learning,
  title = {Learning Molecular Mixture Property Using Chemistry-Aware Graph Neural Network},
  author = {Zhang, Hengrui and Lai, Tianxing and Chen, Jie and Manthiram, Arumugam and Rondinelli, James M. and Chen, Wei},
  journal = {PRX Energy},
  volume = {3},
  issue = {2},
  pages = {023006},
  year = {2024},
  publisher = {American Physical Society},
}

@article{xu2025autonomous,
	title = {Autonomous multi-robot synthesis and optimization of metal halide perovskite nanocrystals},
	volume = {16},
	issn = {2041-1723},
	number = {1},
	journal = {Nature Communications},
	author = {Xu, Jinge and Moran, Christopher H. J. and Ghorai, Arup and Bateni, Fazel and Bennett, Jeffrey A. and Mukhin, Nikolai and Latif, Koray and Cahn, Andrew and Jha, Pragyan and Licona, Fernando Delgado and Sadeghi, Sina and Politi, Lior and Abolhasani, Milad},
	year = {2025},
	pages = {7841},
}

@Article{hickman2025atlas,
    author ="Hickman, Riley J. and Sim, Malcolm and Pablo-García, Sergio and Tom, Gary and Woolhouse, Ivan and Hao, Han and Bao, Zeqing and Bannigan, Pauric and Allen, Christine and Aldeghi, Matteo and Aspuru-Guzik, Alán",
    title  ="Atlas: a brain for self-driving laboratories",
    journal  ="Digital Discovery",
    year  ="2025",
    volume  ="4",
    issue  ="4",
    pages  ="1006-1029",
    publisher  ="RSC",
}

@article{lin2025visual,
	title = {A {Visual} {Dataset} for {Anomaly} {Detection} in {Self}-{Driving} {Laboratories}},
    author = {Lin, Shiwei and Chen, Zesen and Jia, Xiaobin and Wang, Yi and Wang, Chenxu and Zhang, Shuyuan and Ding, Xiaozhen and Du, Boyuan and Ding, Weili and Liu, Huaping},
	volume = {12},
	issn = {2052-4463},
	number = {1},
	journal = {Scientific Data},
	year = {2025},
	pages = {1787},
}

@InProceedings{khaing2019transparent,
    author="Khaing, May Phyo and Masayuki, Mukunoki",
    editor="Zin, Thi Thi
    and Lin, Jerry Chun-Wei",
    title="Transparent Object Detection Using Convolutional Neural Network",
    booktitle="Big Data Analysis and Deep Learning Applications",
    year="2019",
    publisher="Springer Singapore",
    address="Singapore",
    pages="86--93",
    isbn="978-981-13-0869-7"
}

@InProceedings{wang2024novel,
    author="Wang, Hanfang and Chai, Yi and Tao, Tao",
    editor="S. Shmaliy, Yuriy",
    title="A Novel Transparent Object Detection Approach Based on Boundary Optimization",
    booktitle="8th International Conference on Computing, Control and Industrial Engineering (CCIE2024)",
    year="2024",
    publisher="Springer Nature Singapore",
    address="Singapore",
    pages="223--233",
    isbn="978-981-97-6937-7"
}

@article{guohua2019transparent,
    year = {2019},
    publisher = {IOP Publishing},
    volume = {1183},
    number = {1},
    pages = {012011},
    author = {Guo-Hua, Chen and Jun-Yi, Wang and Ai-Jun, Zhang},
    title = {Transparent object detection and location based on RGB-D camera},
    journal = {Journal of Physics: Conference Series},
}

@article{ruoning2024transparent,
    title = {Transparent objects segmentation based on polarization imaging and deep learning},
    journal = {Optics Communications},
    volume = {555},
    pages = {130246},
    year = {2024},
    issn = {0030-4018},
    author = {Ruoning Yu and Wenyi Ren and Man Zhao and Jian Wang and Dan Wu and Yingge Xie},
}

@article{zheng2022glassnet,
    author = {Zheng, Chengyu and Shi, Ding and Yan, Xuefeng and Liang, Dong and Wei, Mingqiang and Yang, Xin and Guo, Yanwen and Xie, Haoran},
    title = {GlassNet: Label Decoupling-based Three-stream Neural Network for Robust Image Glass Detection},
    journal = {Computer Graphics Forum},
    volume = {41},
    number = {1},
    pages = {377-388},
    year = {2022}
}

@inproceedings{kirillov2023segment,
  title={Segment anything},
  author={Kirillov, Alexander and Mintun, Eric and Ravi, Nikhila and Mao, Hanzi and Rolland, Chloe and Gustafson, Laura and Xiao, Tete and Whitehead, Spencer and Berg, Alexander C and Lo, Wan-Yen and others},
  booktitle={Proceedings of the IEEE/CVF international conference on computer vision},
  pages={4015--4026},
  year={2023}
}

@article{macleod2020selfdriving,
    author = {B. P. MacLeod  and F. G. L. Parlane  and T. D. Morrissey  and F. Häse  and L. M. Roch  and K. E. Dettelbach  and R. Moreira  and L. P. E. Yunker  and M. B. Rooney  and J. R. Deeth  and V. Lai  and G. J. Ng  and H. Situ  and R. H. Zhang  and M. S. Elliott  and T. H. Haley  and D. J. Dvorak  and A. Aspuru-Guzik  and J. E. Hein  and C. P. Berlinguette },
    title = {Self-driving laboratory for accelerated discovery of thin-film materials},
    journal = {Science Advances},
    volume = {6},
    number = {20},
    pages = {eaaz8867},
    year = {2020},
}

@misc{sam3dteam2025sam3d3dfyimages,
      title={SAM 3D: 3Dfy Anything in Images}, 
      author={SAM 3D Team and Xingyu Chen and Fu-Jen Chu and Pierre Gleize and Kevin J Liang and Alexander Sax and Hao Tang and Weiyao Wang and Michelle Guo and Thibaut Hardin and Xiang Li and Aohan Lin and Jiawei Liu and Ziqi Ma and Anushka Sagar and Bowen Song and Xiaodong Wang and Jianing Yang and Bowen Zhang and Piotr Dollár and Georgia Gkioxari and Matt Feiszli and Jitendra Malik},
      year={2025},
      eprint={2511.16624},
      archivePrefix={arXiv},
      primaryClass={cs.CV},
      url={https://arxiv.org/abs/2511.16624}, 
}

@article{mathai2022transparent,
	title = {Transparent {Object} {Reconstruction} {Based} on {Compressive} {Sensing} and {Super}-{Resolution} {Convolutional} {Neural} {Network}},
	volume = {12},
	issn = {2190-7439},
	number = {4},
	journal = {Photonic Sensors},
	author = {Mathai, Anumol and Mengdi, Li and Lau, Stephen and Guo, Ningqun and Wang, Xin},
	month = apr,
	year = {2022},
	pages = {220413},
}

@misc{redmon2016lookonceunifiedrealtime,
      title={You Only Look Once: Unified, Real-Time Object Detection}, 
      author={Joseph Redmon and Santosh Divvala and Ross Girshick and Ali Farhadi},
      year={2016},
      eprint={1506.02640},
      archivePrefix={arXiv},
      primaryClass={cs.CV},
      url={https://arxiv.org/abs/1506.02640}, 
}

@article{rani2022object,
	title = {Object detection and recognition using contour based edge detection and fast {R}-{CNN}},
	volume = {81},
	issn = {1573-7721},
	number = {29},
	journal = {Multimedia Tools and Applications},
	author = {Rani, Shilpa and Ghai, Deepika and Kumar, Sandeep},
	month = dec,
	year = {2022},
	pages = {42183--42207},
}

@article{wang2022detection,
  title={Detection of glass insulators using deep neural networks based on optical imaging},
  author={Wang, Jinyu and Li, Yingna and Chen, Wenxiang},
  journal={Remote Sensing},
  volume={14},
  number={20},
  pages={5153},
  year={2022},
  publisher={MDPI}
}

@article{abolhasani2023rise,
	title = {The rise of self-driving labs in chemical and materials sciences},
	volume = {2},
	issn = {2731-0582},
	number = {6},
	journal = {Nature Synthesis},
	author = {Abolhasani, Milad and Kumacheva, Eugenia},
	year = {2023},
	pages = {483--492},
}

@article{tom2024selfdriving,
    author = {Tom, Gary and Schmid, Stefan P. and Baird, Sterling G. and Cao, Yang and Darvish, Kourosh and Hao, Han and Lo, Stanley and Pablo-García, Sergio and Rajaonson, Ella M. and Skreta, Marta and Yoshikawa, Naruki and Corapi, Samantha and Akkoc, Gun Deniz and Strieth-Kalthoff, Felix and Seifrid, Martin and Aspuru-Guzik, Alán},
    title = {Self-Driving Laboratories for Chemistry and Materials Science},
    journal = {Chemical Reviews},
    volume = {124},
    number = {16},
    pages = {9633-9732},
    year = {2024}
}

@article{tobias2025autonomous,
    author = {Tobias, Alexander V.  and Wahab, Adam },
    title = {Autonomous ‘self-driving’ laboratories: a review of technology and policy implications},
    journal = {Royal Society Open Science},
    volume = {12},
    number = {7},
    pages = {250646},
    year = {2025}
}

@article{bayley2024autonomous,
    title = {Autonomous chemistry: Navigating self-driving labs in chemical and material sciences},
    journal = {Matter},
    volume = {7},
    number = {7},
    pages = {2382-2398},
    year = {2024},
    issn = {2590-2385},
    author = {Oliver Bayley and Elia Savino and Aidan Slattery and Timothy Noël},
}

@article{coley2019robotic,
  title={A robotic platform for flow synthesis of organic compounds informed by AI planning},
  author={Coley, Connor W and Thomas III, Dale A and Lummiss, Justin AM and Jaworski, Jonathan N and Breen, Christopher P and Schultz, Victor and Hart, Travis and Fishman, Joshua S and Rogers, Luke and Gao, Hanyu and others},
  journal={Science},
  volume={365},
  number={6453},
  pages={eaax1566},
  year={2019},
  publisher={American Association for the Advancement of Science}
}

@article{burger2020mobile,
  title={A mobile robotic chemist},
  author={Burger, Benjamin and Maffettone, Phillip M and Gusev, Vladimir V and Aitchison, Catherine M and Bai, Yang and Wang, Xiaoyan and Li, Xiaobo and Alston, Ben M and Li, Buyi and Clowes, Rob and others},
  journal={Nature},
  volume={583},
  number={7815},
  pages={237--241},
  year={2020},
  publisher={Nature Publishing Group UK London}
}

@article{fushimi2025development,
	title = {Development of the autonomous lab system to support biotechnology research},
	volume = {15},
	issn = {2045-2322},
	number = {1},
	journal = {Scientific Reports},
	author = {Fushimi, Keiji and Nakai, Yusuke and Nishi, Akiko and Suzuki, Ryo and Ikegami, Masahiro and Nimura, Risa and Tomono, Taichi and Hidese, Ryota and Yasueda, Hisashi and Tagawa, Yusuke and Hasunuma, Tomohisa},
	year = {2025},
	pages = {6648},
}

@article{lysenkov2013recognition,
  title={Recognition and pose estimation of rigid transparent objects with a kinect sensor},
  author={Lysenkov, Ilya and Eruhimov, Victor and Bradski, Gary},
  journal={Robotics},
  volume={273},
  number={273-280},
  pages={2},
  year={2013},
  publisher={MIT Press}
}

@article{yan2021mirrornet,
  title={Mirrornet: Bio-inspired camouflaged object segmentation},
  author={Yan, Jinnan and Le, Trung-Nghia and Nguyen, Khanh-Duy and Tran, Minh-Triet and Do, Thanh-Toan and Nguyen, Tam V},
  journal={IEEE access},
  volume={9},
  pages={43290--43300},
  year={2021},
  publisher={IEEE}
}

@inproceedings{xie2020segmenting,
  title={Segmenting transparent objects in the wild},
  author={Xie, Enze and Wang, Wenjia and Wang, Wenhai and Ding, Mingyu and Shen, Chunhua and Luo, Ping},
  booktitle={European conference on computer vision},
  pages={696--711},
  year={2020},
  organization={Springer}
}

@article{pulli2025enhancing,
  title={Enhancing Transparent Object Pose Estimation: A Fusion of GDR-Net and Edge Detection},
  author={Pulli, Tessa and H{\"o}nig, Peter and Thalhammer, Stefan and Hirschmanner, Matthias and Vincze, Markus},
  journal={arXiv preprint arXiv:2502.12027},
  pages = {},
  year={2025}
}

@article{canny2009computational,
  title={A computational approach to edge detection},
  author={Canny, John},
  journal={IEEE Transactions on pattern analysis and machine intelligence},
  number={6},
  pages={679--698},
  year={2009},
  publisher={Ieee}
}

@article{bradski2000opencv,
  title={The opencv library.},
  author={Bradski, Gary},
  journal={Dr. Dobb's Journal: Software Tools for the Professional Programmer},
  volume={25},
  number={11},
  pages={120--123},
  year={2000},
  publisher={Miller Freeman Inc.}
}

@article{ballard1981generalizing,
  title={Generalizing the Hough transform to detect arbitrary shapes},
  author={Ballard, Dana H},
  journal={Pattern recognition},
  volume={13},
  number={2},
  pages={111--122},
  year={1981},
  publisher={Elsevier}
}

@software{shapely,
  title        = {Shapely},
  author       = {Gillies, Sean and van der Wel, Casper and Van den Bossche, Joris and Taves, Mike W. and Arnott, Joshua and Ward, Brendan C. and others},
  year         = {2025},
  version      = {2.1.1},
  date         = {2025-05-19},
  doi          = {10.5281/zenodo.5597138},
  url          = {https://github.com/shapely/shapely},
  license      = {BSD-3-Clause}
}

@article{paszke2019pytorch,
  title={Pytorch: An imperative style, high-performance deep learning library},
  author={Paszke, Adam and Gross, Sam and Massa, Francisco and Lerer, Adam and Bradbury, James and Chanan, Gregory and Killeen, Trevor and Lin, Zeming and Gimelshein, Natalia and Antiga, Luca and others},
  journal={Advances in neural information processing systems},
  volume={32},
  pages = {},
  year={2019}
}

@book{venables2000introduction,
  title={Introduction to surface and thin film processes},
  author={Venables, John},
  year={2000},
  publisher={Cambridge university press}
}

@inproceedings{lai2015transparent,
  title={Transparent object detection using regions with convolutional neural network},
  author={Lai, Po-Jen and Fuh, Chiou-Shann},
  booktitle={IPPR conference on computer vision, graphics, and image processing},
  volume={2},
  year={2015}
}

@article{han2023segment,
  title={Segment anything model (sam) meets glass: Mirror and transparent objects cannot be easily detected},
  author={Han, Dongsheng and Zhang, Chaoning and Qiao, Yu and Qamar, Maryam and Jung, Yuna and Lee, SeungKyu and Bae, Sung-Ho and Hong, Choong Seon},
  journal={arXiv preprint arXiv:2305.00278},
  pages = {},
  year={2023}
}

@inproceedings{wang2021gdr,
  title={Gdr-net: Geometry-guided direct regression network for monocular 6d object pose estimation},
  author={Wang, Gu and Manhardt, Fabian and Tombari, Federico and Ji, Xiangyang},
  booktitle={Proceedings of the IEEE/CVF conference on computer vision and pattern recognition},
  pages={16611--16621},
  year={2021}
}

@article{ge2021yolox,
  title={Yolox: Exceeding yolo series in 2021},
  author={Ge, Zheng and Liu, Songtao and Wang, Feng and Li, Zeming and Sun, Jian},
  journal={arXiv preprint arXiv:2107.08430},
  pages = {},
  year={2021}
}
\bibliographystyle{unsrt} %the RSC's .bst file

% \newpage
% \beginsupplement
% \include{supp}

\end{document}